\documentclass{article}
\usepackage[table]{xcolor}

\usepackage[final]{corl_2026} %

\usepackage{booktabs}
\usepackage{makecell}
\usepackage{graphicx}
\usepackage{amssymb}
\usepackage{multirow}
\usepackage{threeparttable}

\newcommand{\biroad}{\textsc{BiRoAD}}
\newcommand{\twoval}[2]{#1\,/\,#2}

\title{BiRoAD: Learning Shared and Role-Adaptive Representations for Bimanual Manipulation}

\author{
  Yan Shen$^{1,4 *}$ \quad
  Yuchen Liu$^{1 *}$ \quad
  Feng Jiang$^{1}$ \quad
  Hangtian Hu$^{1}$ \\
  \bf Xiaoqi Li$^{1,4}$ \quad
  Shu Chen$^{3}$ \quad
  Ruihai Wu$^{2}$ \quad
  Hao Dong$^{1,4,\dagger}$ \\[0.5em]
  $^{1}$CFCS, School of Computer Science, Peking University \quad
  $^{2}$UC Berkeley \\
  $^{3}$Hong Kong University of Science and Technology (Guangzhou) \quad
  $^{4}$PrimeBot Research Institute \\ [0.3em]
  $^{*}$Equal contribution. \quad
  $^{\dagger}$Corresponding author.
}

\begin{document}
\maketitle

\begin{abstract}

Bimanual manipulation requires policies that coordinate two arms while adapting their functional roles to scene geometry, object configuration, and task context. Learning such scene-conditioned role adaptation remains challenging, as demonstrations may contain uneven role distributions that limit generalization to underrepresented arm--role configurations. In addition, many bimanual policies predict actions in fixed left- and right-arm action spaces. While this provides a natural parameterization for robot control, it does not explicitly specify how behaviors should transform when functional roles are exchanged across arms. Across different scene initializations, the two arms may follow a similar coordination pattern, but the role-specific behavior assigned to each arm should change with the scene.
Therefore, we propose \textbf{BiRoAD}, a \textbf{Bimanual Role-Adaptive Decomposition} framework for learning shared and role-adaptive representations in bimanual policies. Given bimanual trajectory or action-token features, BiRoAD decomposes these features into swap--symmetric and swap--antisymmetric components: the former captures coordination structure invariant to arm exchange, and the latter captures role-specific distinctions that vary consistently with functional role assignment. The two components are then recomposed as residual updates to the original paired arm representations, allowing BiRoAD to serve as a modular feature transformation without changing the policy inputs, imitation-learning objective, or requiring manually defined role labels. Across multiple bimanual manipulation tasks with balanced and imbalanced role distributions, BiRoAD improves robustness across role configurations over corresponding base policies, with notable gains on underrepresented role configurations. Project Page: \href{https://sxy7147.github.io/biroad-website/}{\nolinkurl{sxy7147.github.io/biroad-website/}}.

\end{abstract}

\vspace{-2mm}
\keywords{Bimanual Manipulation, Imitation Learning}

\section{Introduction}
\label{sec:intro}

\begin{figure}[t]
\centering
\includegraphics[width=\linewidth]{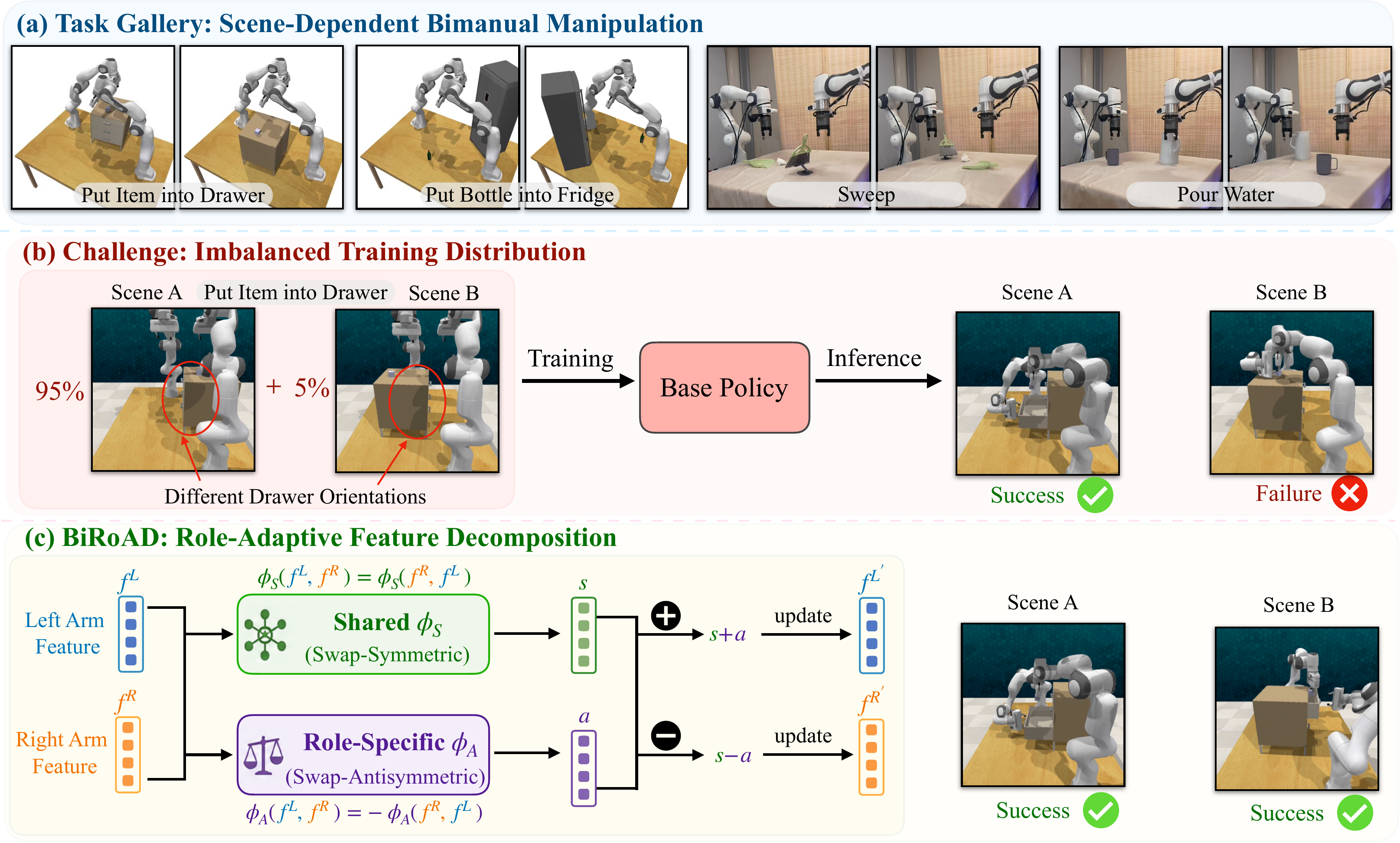}
\vspace{-5mm}
\caption{
\textbf{Overview of BiRoAD.}
(a) Bimanual tasks in which scene initializations induce different functional roles for two arms.
(b) Under imbalanced training distributions, a base policy may favor frequent arm--role configurations and perform less reliably when the scene induces a less common role assignment.
(c) BiRoAD decomposes paired arm features into swap-symmetric shared coordination and swap-antisymmetric role-specific components, then recomposes them as residual updates to encourage structured cross-arm sharing while preserving scene-dependent role adaptation.
}
\vspace{-2mm}
\label{fig:teaser}
\end{figure}

Bimanual manipulation enables robots to coordinate two arms when interacting with objects and environments, extending their capabilities beyond single-arm operation~\cite{grannen2023stabilize,mu2025robotwin,zhao2023learning,grotz2024peract2,fu2024mobile,chen2025robotwin}. In many manipulation tasks, variations in object configuration and workspace constraints require the two arms to adapt their functional roles to the scene~\cite{liu2024voxact,gao2024bi}. Depending on the scene, either arm may need to take a leading, supporting, or operating role. However, demonstrations may exhibit uneven role distributions, where certain arm--role configurations appear more frequently than their mirrored or less common counterparts. Such imbalance can bias learned policies toward frequent arm-specific behaviors and hinder generalization to underrepresented role configurations. This creates a central challenge for policy learning: a bimanual policy should capture coordination structure shared across the arm pair while preserving role-specific distinctions that adapt to the scene.

Recent work has made substantial progress in bimanual manipulation~\cite{zhao2023learning,gkanatsios20253d,lu2025anybimanual,black2024pi_0,liu2025rdt,intelligence2025pi_}, with many policies predicting bimanual actions in fixed left- and right-arm action spaces as a natural and effective parameterization for robot control. Although modern architectures can support cross-arm information sharing through joint representations, they do not explicitly specify how behaviors should transform when functional roles are exchanged across arms. This ambiguity can be amplified under uneven role distributions, where frequently observed behaviors may become coupled with arm identity rather than grounded in the scene-conditioned functional role. These observations motivate representations that promote structured cross-arm sharing while retaining role-specific distinctions for scene-dependent adaptation.

In this work, as shown in Figure~\ref{fig:teaser}, we propose \textbf{BiRoAD}, a \textbf{Bi}manual \textbf{Ro}le-\textbf{A}daptive \textbf{D}ecomposition framework for learning shared and role-adaptive representations in bimanual policies. BiRoAD is motivated by the requirements of role-adaptive bimanual collaboration, which involves two complementary structures: shared coordination patterns that should remain consistent under arm exchange, and role-specific variation that should change consistently with the assigned functional roles. Given paired arm trajectory or action-token features, BiRoAD applies a symmetric--antisymmetric decomposition to organize bimanual interaction into these components. The swap--symmetric component models coordination structure that is invariant to arm exchange, while the swap--antisymmetric component captures role-specific distinctions that vary with the assigned roles. The outputs of the two branches are then recomposed as residual updates to the original paired arm representations, allowing BiRoAD to serve as a modular feature transformation without changing the policy inputs or imitation-learning objective. BiRoAD does not require additional annotations or manually defined role labels, which are often task-dependent and difficult to define consistently across diverse bimanual behaviors. Overall, BiRoAD provides a local symmetry-aware inductive bias that encourages both cross-arm sharing and role-specific differentiation within bimanual representations.

To evaluate role adaptation across varied scene initializations, we introduce a bimanual manipulation evaluation suite built on PerAct2~\cite{grotz2024peract2}, where different initializations induce different functional role assignments within the same task. The suite covers diverse bimanual coordination patterns and supports both balanced and imbalanced role distributions. Beyond average task success, we emphasize role-balanced robustness, which assesses whether a policy can adapt arm roles according to the scene rather than relying primarily on frequent arm-specific behaviors. Across these tasks, BiRoAD improves role-balanced robustness over corresponding base policies, with particularly clear gains on underrepresented and weaker-performing configurations. Together with real-world validation, these results suggest that bimanual policy learning benefits from explicitly structuring representations into shared coordination structure and role-specific distinctions.
To summarize, in this paper:

\begin{itemize}
\vspace{-1mm}
\item We study scene-conditioned role adaptation in bimanual manipulation, where varied scene initializations require policies to adjust functional role assignments across the two arms, especially under imbalanced role distributions.

\item We propose BiRoAD, a modular symmetric--antisymmetric decomposition framework for learning shared and role-adaptive bimanual representations. BiRoAD can be integrated into existing policies as a residual feature transformation without requiring manual role labels.

\item We introduce a bimanual manipulation evaluation suite with varied object initializations and balanced or imbalanced role distributions, and show that BiRoAD improves role-balanced robustness over corresponding base policies.
\end{itemize}

\vspace{-1mm}
\section{Related Work}
\label{sec:related}
\vspace{-1mm}

\subsection{Bimanual Manipulation}
\vspace{-1mm}

Bimanual manipulation expands robot capabilities by requiring spatial and temporal coordination between two arms~\cite{grannen2023stabilize, zhao2023learning, fu2024mobile, aldaco2024aloha, grannen2022learning, ren2024enabling, zhao2022dualafford, shen2025biassemble}.
Recent progress has been driven by teleoperation systems, benchmarks, and learning-based policies.
ACT~\cite{zhao2023learning} uses temporally chunked action prediction for fine-grained bimanual manipulation, while 3D FlowMatch Actor~\cite{gkanatsios20253d} predicts bimanual trajectories from 3D scene features using flow matching.
Benchmarks~\cite{chen2025robotwin, wang2025roboeval, peng2026bicoord, jiang2026robowm} such as PerAct2~\cite{grotz2024peract2} and RoboTwin~\cite{mu2025robotwin} support evaluation across diverse bimanual tasks, scene configurations, and coordination requirements.
Other studies have explored bimanual manipulation settings in which two arms assume complementary functional roles~\cite{shen2026bipremanip, zhou2025you, zhou2025vlbiman, palma2026bimanual, jiang2025rethinking, liu2025factr, lee2024interact, liang2026a3d, lu2025anybimanual, yu2025manigaussian++}.
VoxAct-B~\cite{liu2024voxact} considers scene-dependent acting and stabilizing roles, while EquiBim~\cite{zhang2026equibim} enforces bilateral equivariance between symmetrically transformed observations and actions.
Complementary to these approaches, BiRoAD does not require explicit role labels, predefined functional roles, or hard policy-level symmetry constraints. Instead, it studies how internal paired-arm representations can support scene-conditioned role adaptation across varying role configurations.

\vspace{-1mm}
\subsection{Imitation Learning for Robot Manipulation}
\vspace{-1mm}

Imitation learning enables robot policies to acquire manipulation skills from demonstrations, with recent progress in generative policy learning and vision-language-action policies~\cite{lipman2022flow, ke20243d, qu2025spatialvla, li2025manidp, yan2025maniflow, chi2025diffusion, ze20243d, feng2025vidar, brohan2022rt, li20253ds, li2025object, li2024manipllm}.
Representative systems include ACT~\cite{zhao2023learning}, which uses temporally chunked action prediction, and 3D FlowMatch Actor~\cite{gkanatsios20253d}, which conditions flow-matching policies on 3D scene features.
RDT-1B~\cite{liu2025rdt} scales diffusion-based imitation learning to bimanual manipulation with a large Robotics Diffusion Transformer and a unified action space across robot embodiments.
Recent vision-language-action policies such as $\pi_0$~\cite{black2024pi_0} and $\pi_{0.5}$~\cite{intelligence2025pi_} further scale manipulation learning across diverse tasks and embodiments.
BiRoAD further introduces a lightweight feature-level module that applies role-adaptive decomposition to paired arm features, supporting scene-conditioned role adaptation without changing the policy interface or imitation-learning objectives.

\vspace{-1mm}
\section{Problem Formulation}
\label{sec:formulation}
\vspace{-1mm}
We consider imitation learning for bimanual manipulation from a dataset \(\mathcal{D}=\{(\tau_i,l_i)\}_{i=1}^{N}\) of \(N\) expert demonstrations, where each demonstration consists of a trajectory \(\tau_i\) and a task description \(l_i\). We represent each trajectory as \(\tau_i=\{(o_t,s_t,a_t)\}_{t=1}^{T_i}\), where \(o_t\) denotes the scene observation, \(s_t=(s_t^L,s_t^R)\) denotes the proprioceptive states of the left and right arms, and \(a_t=(a_t^L,a_t^R)\) denotes their corresponding actions.

In \(\mathcal{D}\), different scene initializations may induce different functional role assignments between the two arms. For a given task family, let \(\rho_L\) and \(\rho_R\) denote the frequencies with which the left or right arm assumes a task-leading or operating role, respectively. In practice, these frequencies can be imbalanced, \emph{i.e.}, \(\rho_L \neq \rho_R\), when some arm--role configurations occur more often than their mirrored or less common counterparts in the demonstration data. In our setting, different initializations of the same task share the same language instruction \(l\), which specifies the task goal, such as ``placing an object into a target drawer,'' but does not specify the scene-dependent role assignment, such as which arm should open the drawer and which should grasp the object. The policy must therefore infer the appropriate role assignment from the scene observation when predicting paired arm actions, without relying on manual role annotations or additional role-assignment supervision.

\vspace{-1mm}
\section{Method}
\label{sec:method}
\vspace{-1mm}

\subsection{Overview}
\label{subsec:overview}
\begin{figure}[t]
\centering
\includegraphics[width=\linewidth]{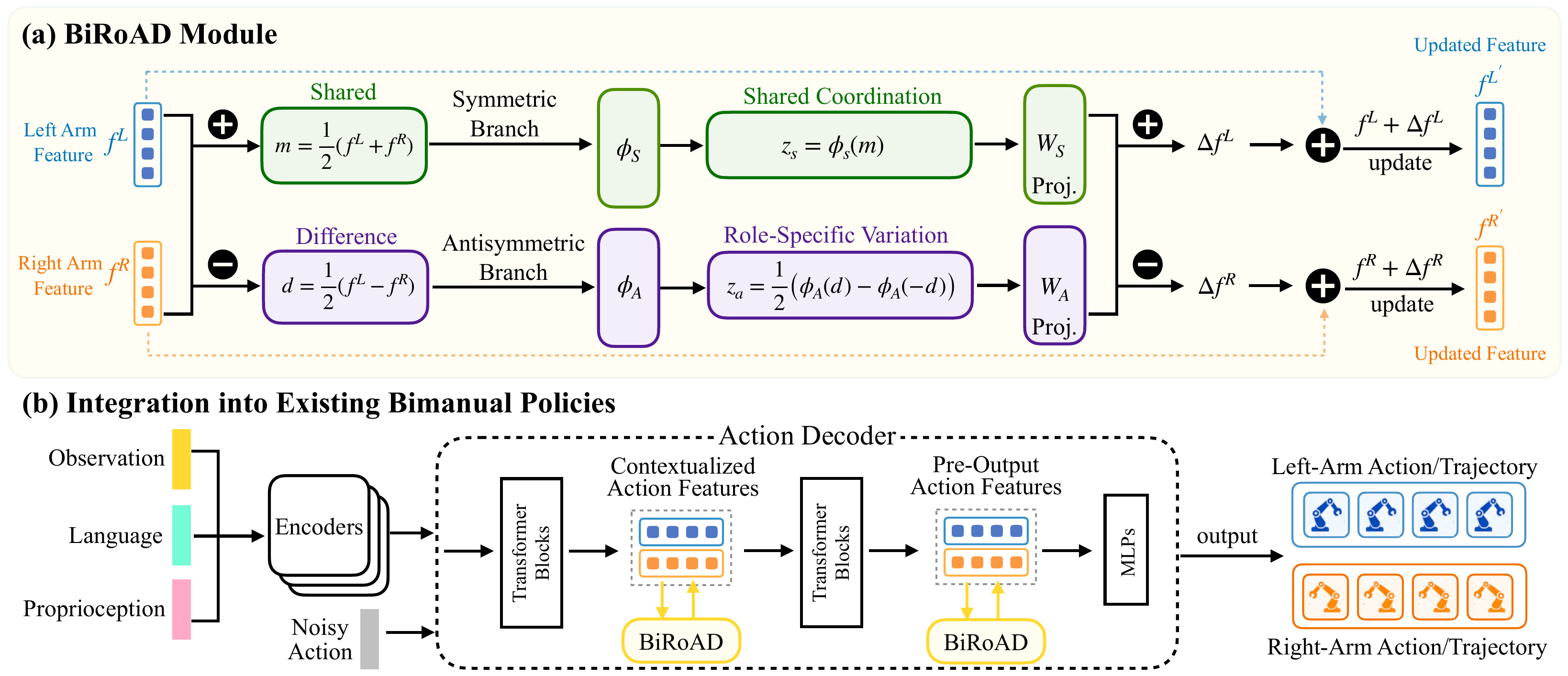}
\vspace{-5mm}
\caption{
\textbf{BiRoAD module and policy integration.}
(a) BiRoAD performs symmetric--antisymmetric feature decomposition and recomposes the resulting components as residual updates.
(b) BiRoAD refines paired arm trajectory or action-token features at selected points in the action-generation module without changing the policy interface.
}
\vspace{-1mm}
\label{fig:pipeline}
\end{figure}

\vspace{-1mm}
As illustrated in Figure~\ref{fig:pipeline}, we propose BiRoAD, a modular feature decomposition method for bimanual policies. BiRoAD operates on paired left- and right-arm action representations within the policy's action-generation pathway. Given paired arm features \(F=(f^L, f^R)\), BiRoAD decomposes them into two complementary components: a swap-invariant component \(S\), satisfying \(S(f^L, f^R)=S(f^R, f^L)\), which captures coordination structure shared across the two arms; and a swap-antisymmetric component \(A\), satisfying \(A(f^L, f^R)=-A(f^R, f^L)\), which captures relative, role-specific distinctions between the arms. The concrete construction is described in Section~\ref{subsec:sym_anti}. BiRoAD then recomposes these components as residual updates to the original arm representations, preserving the policy interface while encouraging the action representation to model both shared bimanual coordination and scene-dependent role variation. Therefore, BiRoAD can be integrated into existing policy architectures without changing their input--output interfaces or requiring additional role annotations. Representative integrations are discussed in Section~\ref{subsec:integration}.

\vspace{-1mm}
\subsection{Symmetric--Antisymmetric Role Decomposition}
\label{subsec:sym_anti}
\vspace{-1mm}

BiRoAD is designed to separate shared bimanual coordination from role-specific left--right distinctions within paired arm representations. We instantiate this idea as a residual feature transformation in the action-generation pathway.
Given the left- and right-arm trajectory or action-token features \(f^L\) and \(f^R\) at a selected insertion point in the action-generation module,
BiRoAD first maps the paired features into exchange-symmetric and exchange-antisymmetric coordinates:
\setlength{\abovedisplayskip}{3pt}
\setlength{\belowdisplayskip}{3pt}
\setlength{\abovedisplayshortskip}{3pt}
\setlength{\belowdisplayshortskip}{3pt}
\begin{equation}
    m = \frac{1}{2}(f^L + f^R),
    \qquad
    d = \frac{1}{2}(f^L - f^R).
\end{equation}
The mean component \(m\) is exchange-invariant and captures information shared across the two arms, while the relative component \(d\) changes sign when the order of the two arms is reversed and therefore encodes their ordered left--right difference. The factors \(1/2\) keep the transformed features on a scale comparable to the original representations and yield the inverse relation \(f^L=m+d\), \(f^R=m-d\).

BiRoAD then processes these coordinates with two branches. The symmetric branch is defined as
\setlength{\abovedisplayskip}{3pt}
\setlength{\belowdisplayskip}{3pt}
\setlength{\abovedisplayshortskip}{3pt}
\setlength{\belowdisplayshortskip}{3pt}
\begin{equation}
    z_S = \phi_S(m),
\end{equation}
where \(\phi_S\) is implemented as an MLP. Since \(m\) is exchange-invariant, \(\phi_S\) is also invariant to arm exchange, providing a pathway for modeling coordination structure shared across the arm pair.

The antisymmetric branch models directional differences between the two arm representations by applying an odd transformation to the relative feature:
\setlength{\abovedisplayskip}{3pt}
\setlength{\belowdisplayskip}{3pt}
\setlength{\abovedisplayshortskip}{3pt}
\setlength{\belowdisplayshortskip}{3pt}
\begin{equation}
    z_A = \frac{1}{2}\left(\phi_A(d) - \phi_A(-d)\right),
\end{equation}
where \(\phi_A\) is implemented as an MLP. This formulation does not require \(\phi_A\) itself to satisfy any architectural symmetry constraint. Instead, antisymmetry is enforced by construction: reversing the ordered comparison maps \(d\) to \(-d\), and therefore maps \(z_A\) to \(-z_A\). The factor \(1/2\) keeps the output scale comparable to the symmetric branch, although the antisymmetric property would hold without it. This branch provides a structured pathway for modeling role-specific differences that depend on the direction of the left--right comparison.

The two components are recomposed into arm-specific residual updates, with the symmetric component shared across arms and the antisymmetric component added with opposite signs:
\setlength{\abovedisplayskip}{3pt}
\setlength{\belowdisplayskip}{3pt}
\setlength{\abovedisplayshortskip}{3pt}
\setlength{\belowdisplayshortskip}{3pt}
\begin{equation}
    \Delta f^L = W_S z_S + W_A z_A,
    \qquad
    \Delta f^R = W_S z_S - W_A z_A,
\end{equation}
where \(W_S\) and \(W_A\) are learnable projections. The sign pattern mirrors the symmetric--antisymmetric coordinates: \(W_S z_S\) contributes a common residual to the arm pair, while \(W_A z_A\) modulates their relative left--right variation and therefore enters the two residuals with opposite signs. The updated features are obtained by residual addition:
\setlength{\abovedisplayskip}{3pt}
\setlength{\belowdisplayskip}{3pt}
\setlength{\abovedisplayshortskip}{3pt}
\setlength{\belowdisplayshortskip}{3pt}
\begin{equation}
    {f^L}' = f^L + \Delta f^L,
    \qquad
    {f^R}' = f^R + \Delta f^R.
\end{equation}
This residual formulation preserves the policy interface while incorporating both shared coordination structure and role-specific differences.

Overall, BiRoAD introduces a local exchange-consistent inductive bias within paired bimanual representations. It does not require the full policy to be strictly equivariant to arm exchange, since the architecture may still include ordered observations, fixed arm embeddings, or separate left- and right-arm action heads. Instead, BiRoAD applies this structure only to selected features in the action-generation pathway, making it a lightweight residual transformation that can be integrated into existing policies without changing their input--output interface or requiring role annotations.

\subsection{Integration into Existing Bimanual Policies}
\label{subsec:integration}
\vspace{-1mm}

We next describe how BiRoAD is integrated into the action-generation pathway of existing bimanual policy backbones. We instantiate the module in two representative policy families: a 3D flow-matching actor~\cite{gkanatsios20253d} and a \(\pi_0\)-style vision-language-action policy~\cite{black2024pi_0}.

\vspace{-2mm}
\paragraph{3D flow-matching actor.}
The 3D flow-matching actor predicts bimanual trajectories by iteratively refining noisy action trajectories conditioned on language instructions and 3D scene representations. Its trajectory decoder progressively transforms task intent, spatial context, and denoising information into arm-specific trajectory features, making it a suitable location for BiRoAD. We therefore apply BiRoAD directly to the corresponding left- and right-arm trajectory features at selected decoder stages, while leaving the surrounding decoder structure unchanged.

Concretely, we insert BiRoAD at three stages of the trajectory decoder: after the initial language and observation conditioning, after trajectory--scene feature interaction, and immediately before the final action prediction heads. These insertion points expose trajectory features at different levels of contextualization, enabling the residual symmetric--antisymmetric update to refine both shared coordination structure and role-specific distinctions during trajectory generation. Other components of the policy, including the encoders, action heads, and flow-matching objective, are kept unchanged.

\vspace{-2mm}
\paragraph{\(\pi_0\)-style VLA policy.}
The \(\pi_0\)-style VLA policy factorizes action generation into a vision-language prefix and an action-generation suffix. The prefix encodes image and language observations into contextual representations, while the suffix combines this context with proprioceptive state, noisy action inputs, and denoising-time information to generate actions. We attach BiRoAD to the suffix rather than the prefix, since the suffix is where action representations are constructed and refined.

Within the suffix, we insert BiRoAD at two stages. First, we apply BiRoAD to the suffix hidden states after they have incorporated the vision-language context, proprioceptive state, timestep, and noisy action information. These hidden states serve as task- and observation-conditioned intermediate action representations, enabling the symmetric--antisymmetric decomposition to operate after the main conditioning signals have been fused. Second, in the output head, we apply BiRoAD to the per-arm latent features derived from the action-horizon hidden states before projecting them to the final action prediction. The vision-language pathway, action normalization, output convention, and training loss are kept unchanged.

\vspace{-2mm}
\paragraph{Training.}
BiRoAD is trained jointly with the base policy using the original imitation-learning or flow-matching objective of the corresponding backbone. We do not introduce additional role labels, role-classification losses, or auxiliary supervision. The parameters of BiRoAD are optimized end-to-end through the same action prediction loss used by the underlying policy.

\vspace{-1mm}
\section{Experiments}
\label{sec:exp}

\vspace{-1mm}
\subsection{Experimental Settings}
\vspace{-1mm}

\paragraph{Simulation Setup and Tasks.}
We build our simulation suite on RLBench~\cite{james2020rlbench}, following the bimanual setup of PerAct2~\cite{grotz2024peract2} with two Franka Emika Panda arms.
To evaluate scene-conditioned role adaptation, we adapt a set of PerAct2 bimanual task families into role-configurable settings, where different object layouts and interaction regions induce different functional role assignments between the two arms.
We use eight task families and construct two configurations for each: a base-role configuration and a role-reversed configuration, yielding sixteen task variants in total.
We further increase task randomization by broadening object initialization ranges.
The full task list and implementation details are provided in the supplementary material.
All visual observations are resized to $128 \times 128$ during training and evaluation.
For each task family, we evaluate on $50{+}50$ held-out episodes from the base-role and role-reversed configurations respectively.
For training, we consider two demonstration compositions: a balanced $50{+}50$ setting and a highly imbalanced $95{+}5$ setting, to evaluate role-balanced learning and generalization to underrepresented role configurations.

\vspace{-2mm}
\paragraph{Evaluation metrics.}
To assess whether a policy can adapt to different arm-role configurations, we first report the configuration-level average success rates under the base-role and role-reversed configurations. We further summarize performance with 4 metrics computed at the task level. For a task \(t\), let \(s_t^{\mathrm{base}}\) and \(s_t^{\mathrm{rev}}\) denote its success rates under the two configurations. We compute \emph{Mean}, \emph{HM}, \emph{Worst}, and \emph{Gap} for this task pair, and report the average of each metric over all tasks. Here, \emph{Mean} captures overall success across configurations, \emph{HM} denotes their harmonic mean and emphasizes balanced performance, \emph{Worst} measures the lower success rate; and \emph{Gap} measures the absolute performance disparity between them. Since a small \emph{Gap} may also result from uniformly low success rates, we interpret \emph{Gap} together with the other metrics rather than as a standalone indicator of performance. Detailed metric definitions and formulas are provided in the supplementary material.

\vspace{-1mm}
\subsection{Main Results}
\vspace{-1mm}

\paragraph{Baselines.}
We select representative bimanual manipulation policies as baselines.
For a fair comparison, all methods use the same observation resolution, action representation, and language setting.
(1) \textbf{3DFA}~\cite{gkanatsios20253d}, a strong 3D manipulation policy that predicts bimanual trajectories from 3D scene features using flow matching; (2) \textbf{$\pi_0$}~\cite{black2024pi_0}, a vision-language-action flow policy that generates robot actions through conditional denoising with visual, language, and proprioceptive context; (3) \textbf{ACT}~\cite{zhao2023learning}, a transformer-based imitation-learning policy with temporally chunked action prediction; (4) \textbf{DP3}~\cite{ze20243d}, a diffusion-based visuomotor policy that predicts action trajectories from 3D observations; and (5) \textbf{PerAct2}~\cite{grotz2024peract2}, a voxel-based bimanual manipulation policy built on PerAct~\cite{shridhar2023perceiver}. Following its original protocol, PerAct2 is trained separately for each task (using both the base-role and role-reversed configurations), whereas the other baselines are trained in a multitask setting.

\begin{table*}[t]
\centering
\small
\renewcommand{\arraystretch}{1}
\caption{
Main simulation results under balanced and imbalanced training distributions.
}
\label{tab:main_results}
\resizebox{\textwidth}{!}{
\begin{tabular}{@{}l crrrr | crrrr@{}}
\toprule
\multirow[c]{2}{*}{Method}
& \multicolumn{5}{c}{50:50}
& \multicolumn{5}{c}{95:5} \\
\cmidrule(lr){2-6} \cmidrule(lr){7-11}
& Config.-level Avg. $\uparrow$
& Mean $\uparrow$
& HM $\uparrow$
& Worst $\uparrow$
& Gap $\downarrow$
& Config.-level Avg. $\uparrow$
& Mean $\uparrow$
& HM $\uparrow$
& Worst $\uparrow$
& Gap $\downarrow$ \\
\midrule
DP3
& \twoval{11.75}{12.25}
& 12.00
& 11.46
& 10.25
& 3.50
& \twoval{12.00}{6.00}
& 9.00
& 5.80
& 5.25
& 7.50 \\

ACT
& \twoval{20.25}{19.25}
& 19.75
& 16.89
& 13.50
& 12.50
& \twoval{32.75}{6.25}
& 19.50
& 7.36
& 6.25
& 26.50 \\

PerAct$^{2}$
& \twoval{22.00}{21.25}
& 21.62
& 18.78
& 16.00
& 11.25
& \twoval{13.25}{19.75}
& 16.50
& 13.11
& 12.00
& 9.00 \\

\midrule
\(\pi_0\)
& \twoval{18.75}{24.00}
& 21.38
& 20.64
& 18.25
& 6.25
& \twoval{23.25}{14.00}
& 18.62
& 15.69
& 13.00
& 11.25 \\

\(\pi_0\) + BiRoAD
& \twoval{26.00}{27.25}
& 26.62
& 26.37
& 24.75
& 3.75
& \twoval{36.75}{17.75}
& 27.25
& 21.23
& 17.75
& 19.00 \\

\midrule
3DFA
& \twoval{74.50}{67.50}
& 71.00
& 63.42
& 56.50
& 29.00
& \twoval{81.25}{50.00}
& 65.62
& 56.56
& 46.75
& 37.75 \\

3DFA + BiRoAD
& \textbf{\twoval{83.00}{75.25}}
& \textbf{79.12}
& \textbf{77.41}
& \textbf{71.50}
& 15.25
& \textbf{\twoval{90.50}{63.25}}
& \textbf{76.88}
& \textbf{71.21}
& \textbf{63.25}
& 27.25 \\
\bottomrule
\end{tabular}
}
\end{table*}

\vspace{-2mm}
\paragraph{Analysis.}
Table~\ref{tab:main_results} compares the policies under balanced and highly skewed training distributions. Across both settings, BiRoAD improves the two tested backbones in mean success rate, harmonic mean, and worst-group performance. Under the balanced 50:50 setting, BiRoAD increases the mean success rate for both the \(\pi_0\)-style and 3DFA baselines, suggesting that the symmetric--antisymmetric decomposition can benefit policy learning when both role configurations are equally represented.
The gains are more evident under the 95:5 setting, where policies must generalize to a less frequent role configuration. For 3DFA, BiRoAD improves the configuration-level averages from 81.25/50.00 to 90.50/63.25, with corresponding gains in harmonic mean and worst-group performance. BiRoAD also improves the \(\pi_0\)-style backbone in absolute success rates. Although its Gap increases, both configuration-level averages improve, highlighting that Gap should be interpreted together with absolute success rates rather than in isolation. Overall, these trends suggest that BiRoAD improves role adaptation, especially on weaker or underrepresented configurations. Per-task success rates are provided in the supplementary material.

\vspace{-1mm}
\subsection{Ablation Study}
\label{subsec:ablation}
\vspace{-1mm}

\begin{table}[t]
\centering
\caption{
Ablation study of BiRoAD on 3DFA under the 95:5 training distribution.
}
\vspace{-2mm}
\centering
\small
\setlength{\tabcolsep}{2pt}
\label{tab:ablation}
\resizebox{0.85\textwidth}{!}{
\begin{tabular*}{\linewidth}{@{\extracolsep{\fill}}lccccc@{}}
\toprule
Method
& Config.-level Avg. $\uparrow$
& Mean $\uparrow$
& HM $\uparrow$
& Worst $\uparrow$
& Gap $\downarrow$ \\
\midrule
3DFA (base)
& 81.25\,/\,50.00
& 65.62
& 56.56
& 46.75
& 37.75 \\

+ BiRoAD (direct)
& 82.25\,/\,49.50
& 65.88
& 57.90
& 49.50
& 32.75 \\

+ BiRoAD (early)
& 85.75\,/\,51.25
& 68.50
& 59.40
& 51.00
& 35.00 \\

+ BiRoAD (middle)
& 85.00\,/\,51.75
& 68.38
& 60.75
& 51.75
& 33.25 \\

+ BiRoAD (late)
& 82.25\,/\,\underline{59.75}
& \underline{71.00}
& \underline{65.45}
& \underline{58.00}
& \textbf{26.00} \\

+ BiRoAD (middle+late)
& \underline{86.50}\,/\,55.25
& 70.88
& 63.29
& 55.25
& 31.25 \\

+ Param.-matched MLP (full)
& 86.25\,/\,50.50
& 68.38
& 57.84
& 50.50
& 35.75 \\

+ BiRoAD (full)
& \textbf{90.50\,/\,63.25}
& \textbf{76.88}
& \textbf{71.21}
& \textbf{63.25}
& \underline{27.25} \\
\bottomrule
\end{tabular*}
}
\end{table}

To better understand the design choices in BiRoAD, we conduct ablations under the challenging 95:5 training distribution, where one role configuration is heavily underrepresented. All variants use the same 3DFA backbone, the best-performing backbone in the main comparison. We examine three design factors. First, \textit{direct} replaces the original paired arm features with the BiRoAD features, whereas our default design uses BiRoAD as a residual update. Second, \textit{early}, \textit{middle}, and \textit{late} insert BiRoAD at different trajectory-decoder stages: after initial language and observation conditioning, after trajectory--scene feature interaction, and before the final action prediction heads, respectively. The \textit{middle+late} variant combines the latter two stages, while \textit{full} applies BiRoAD at all three stages. Third, \textit{Param.-matched MLP} uses the same multi-stage insertion pattern as \textit{full}, but replaces the symmetric--antisymmetric decomposition with pairwise MLPs.

Table~\ref{tab:ablation} reports the results. The relatively lower performance of \textit{direct} suggests that BiRoAD is more effective when used as a residual refinement, which preserves backbone representations while introducing role-adaptive structure. Among single-stage variants, \textit{late} achieves the strongest minority-configuration performance, indicating that role-aware coupling is especially useful near action prediction, where arm-specific pose outputs are produced. The \textit{full} model achieves the strongest overall performance, suggesting that early and middle insertions help propagate role-aware features through the decoder, while late insertion shapes the final arm-specific actions. The \textit{Param.-matched MLP} improves the frequent-configuration average but provides smaller gains on the minority configuration, suggesting that the exchange-structured decomposition is important beyond parameter count.

\vspace{-1mm}
\subsection{Real-World Experiments}
\begin{table*}[t]
\centering
\caption{
Real-world success rates under balanced and imbalanced training distributions.
}
\label{tab:real_world}
\setlength{\tabcolsep}{4pt}
\renewcommand{\arraystretch}{1.08}
\resizebox{\textwidth}{!}{
\begin{tabular}{lcccccc|cccccc}
\toprule
\multicolumn{1}{c}{\multirow{2}{*}[-1ex]{Method}}
& \multicolumn{6}{c}{Balanced Setting}
& \multicolumn{6}{c}{Imbalanced Setting} \\
\cmidrule(lr){2-7} \cmidrule(lr){8-13}
& \makecell{Bouquet\\Handover}
& \makecell{Banana\\to Drawer}
& \makecell{Open\\Bottle Cap}
& \makecell{Sweep\\to Dustpan}
& \makecell{Pour\\Water}
& Overall
& \makecell{Bouquet\\Handover}
& \makecell{Banana\\to Drawer}
& \makecell{Open\\Bottle Cap}
& \makecell{Sweep\\to Dustpan}
& \makecell{Pour\\Water}
& Overall \\
\midrule
3DFA
& 0.6 / 0.6
& 0.2 / 0.2
& 0.4 / 0.2
& 0.4 / 0.2
& 0.2 / 0.2
& 0.36 / 0.28
& 0.6 / 0.0
& 0.0 / 0.0
& 0.6 / 0.0
& 0.2 / 0.2
& 0.6 / 0.0
& \textbf{0.40} / 0.04 \\
3DFA + BiRoAD
& 0.6 / 0.6
& 0.4 / 0.2
& 0.4 / 0.4
& 0.2 / 0.2
& 0.4 / 0.2
& \textbf{0.40} / \textbf{0.32}
& 0.6 / 0.2
& 0.4 / 0.0
& 0.4 / 0.2
& 0.2 / 0.0
& 0.2 / 0.2
& 0.36 / \textbf{0.12} \\
\midrule

\(\pi_0\)
& 0.6 / 0.4
& 0.0 / 0.0
& 0.0 / 0.0
& 0.0 / 0.4
& 0.0 / 0.2
& 0.12 / \textbf{0.20}
& 0.2 / 0.4
& 0.0 / 0.0
& 0.0 / 0.0
& 0.0 / 0.2
& 0.0 / 0.2
& 0.04 / 0.16 \\
\(\pi_0\) + BiRoAD
& 0.6 / 0.4
& 0.8 / 0.0
& 0.6 / 0.2
& 0.4 / 0.2
& 0.2 / 0.0
& \textbf{0.52} / 0.16
& 0.2 / 0.6
& 0.0 / 0.4
& 0.2 / 0.2
& 0.2 / 0.2
& 0.0 / 0.2
& \textbf{0.12} / \textbf{0.32} \\
\bottomrule
\end{tabular}
}
\end{table*}

\vspace{-1mm}

\begin{figure}[t]
\centering
\includegraphics[width=\linewidth]{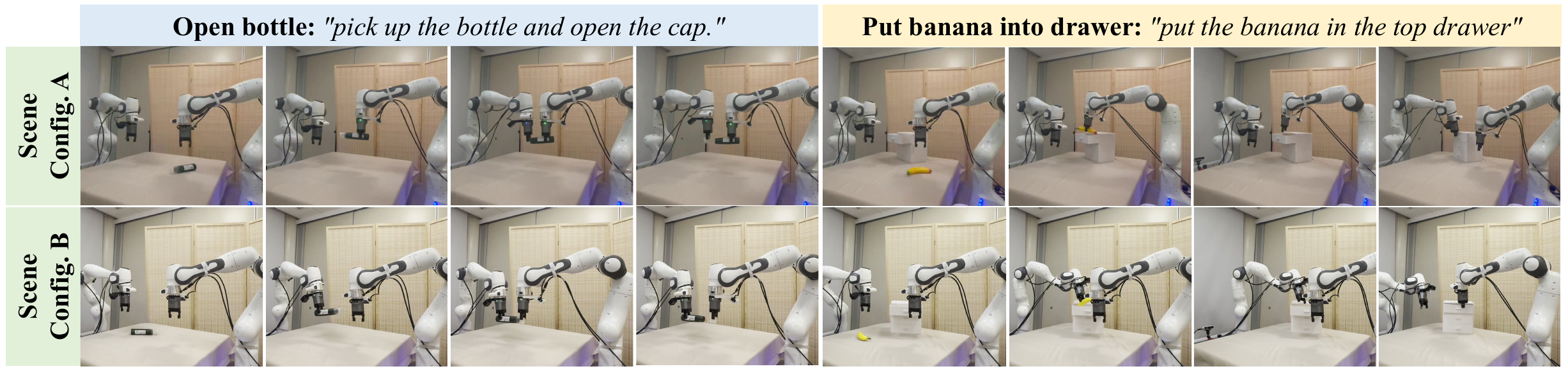}
\vspace{-3mm}
\caption{
\textbf{Visualization of real-world experiments.} Different scene configurations induce different arm-role assignments, requiring the policy to adapt its bimanual behavior to the current scene.
}
\vspace{-1mm}
\label{fig:real_world}
\end{figure}

We further evaluate BiRoAD in a real-world bimanual setup with two Franka Research 3 (FR3) arms equipped with DH-Robotics PGC-series parallel grippers. Observations are captured by one third-person and two wrist-mounted RealSense D405 cameras, and expert demonstrations are collected via human teleoperation. We consider five tasks with two scene configurations that induce different arm-role assignments. In the balanced setting, each method uses 25 demonstrations per configuration. In the imbalanced setting, we use opposite configuration ratios for the two backbone families, 25:5 for 3DFA and 5:25 for \(\pi_0\), to cover both directions of configuration imbalance. This design is motivated by the fact that the two configurations are not fully equivalent in the real setup due to the off-center third-person camera. We therefore focus on within-backbone comparisons between each base policy and its BiRoAD-enhanced counterpart. Each task is evaluated over 5 trials per configuration (10 trials per task), using task success rate as the metric. Figure~\ref{fig:real_world} shows the real-world rollouts, and Table~\ref{tab:real_world} reports the results. Overall, BiRoAD improves both base policies on real-world tasks that require scene-conditioned role adaptation. These trends are consistent with the simulation results.

\vspace{-1mm}
\section{Conclusion}
\label{sec:conclusion}
\vspace{-1mm}

In this work, we studied scene-conditioned role adaptation in bimanual manipulation, where different scene configurations induce different functional roles for the two arms. We proposed BiRoAD, a modular framework that decomposes paired arm representations into swap-symmetric components for shared coordination and swap-antisymmetric components for role-specific variation. Implemented as a residual feature transformation, BiRoAD can be integrated into existing policies without modifying their interfaces, training objectives, or requiring manual role labels. Experimental results in simulation and the real world show that BiRoAD improves role-balanced robustness, particularly on weaker or underrepresented configurations, suggesting that structured shared and role-adaptive representations can benefit bimanual policy learning under scene-dependent role assignments.

\vspace{-2mm}
\paragraph{Limitations.}
\label{sec:limatation}
BiRoAD is a modular representation transformation that complements existing bimanual policies. Its benefits therefore depend on the representational and action-generation capacity of the underlying policy. This method does not directly address broader challenges in manipulation such as perception uncertainty, contact dynamics, or execution variability. Our evaluation focuses on two-arm tasks where scene configurations induce different functional roles between arms. Extending BiRoAD to multi-arm systems and dexterous hands would further test its generality.

\acknowledgments{
This paper is supported by Beijing Natural Science Foundation (26L080330).
}

\clearpage

\bibliography{example}  %

\clearpage
\appendix
\section{Real-World Experiments}

\begin{figure}[h]
\centering
\includegraphics[width=\linewidth]{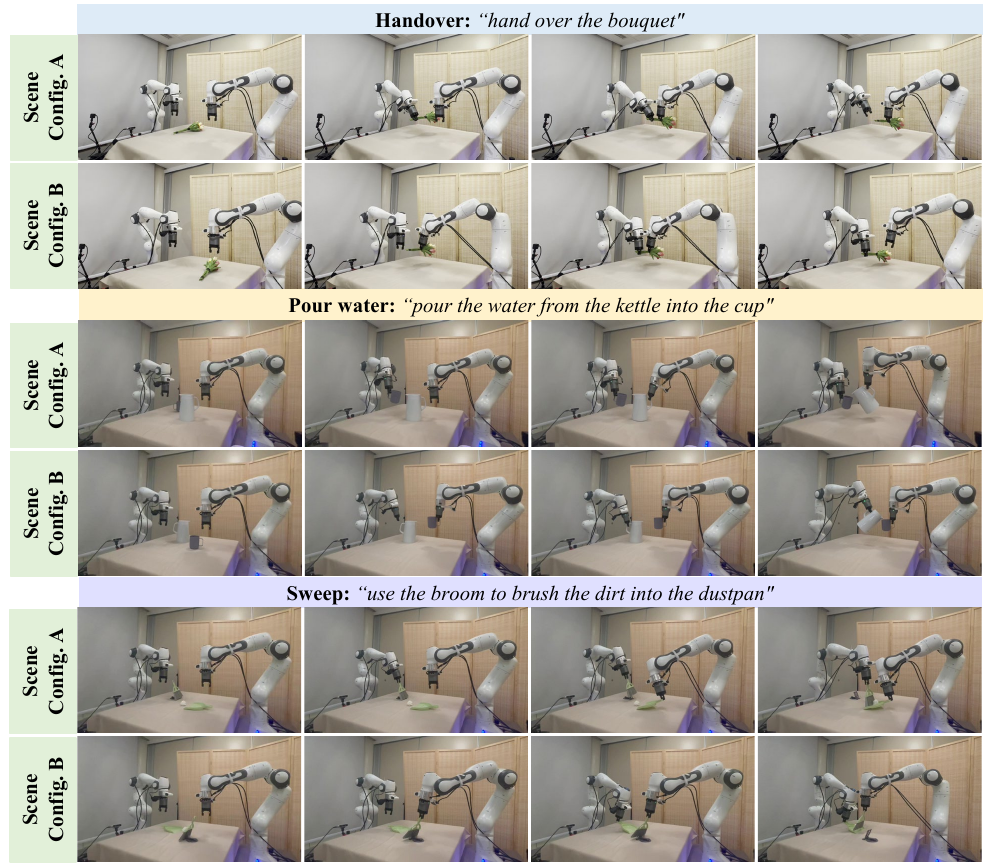}
\caption{
\textbf{Additional real-world experiments.}
We visualize real-world task variants under different scene configurations, which induce different arm-role assignments and require scene-conditioned bimanual behavior.
}
\label{fig:real_data}
\end{figure}

We provide additional descriptions of the real-world tasks used in our experiments.
The benchmark contains five bimanual tasks, each with two scene configurations, yielding a total of ten real-world task variants.
Similar to the simulation benchmark, the two configurations of each task share the same task goal, while variations in object placement, tool placement, or workspace accessibility induce different functional role assignments between the two arms.
Figure~\ref{fig:real_data} shows the exact language instructions used for training and evaluation.

The following paragraphs provide additional clarification on the task procedure and role variations.

\begin{enumerate}
    \item \textbf{Bouquet Handover.}
    One arm grasps the bouquet from the table, while the other arm receives it to complete the handover.
    The bouquet’s location differs between configurations, determining which arm is better suited for the initial grasping role.

    \item \textbf{Put Banana into Drawer.}
    One arm opens the drawer, while the other grasps the banana and places it inside.
    Differences in banana and drawer layouts across configurations require the policy to select which arm should open the drawer and which should manipulate the banana.

    \item \textbf{Open Bottle Cap.}
    One arm picks up the bottle, while the other grasps and removes the cap.
    Variations in bottle placement across configurations affect which arm is better positioned for picking up versus operating roles.

    \item \textbf{Sweep to Dustpan.}
    One arm uses the broom to sweep the dirt, while the other holds the dustpan to collect it.
    The relative positions of the broom, dustpan, and dirt vary between configurations, requiring the policy to assign tool-use roles based on the current scene.

    \item \textbf{Pour Water.}
    One arm grasps and tilts the kettle, while the other stabilizes or positions the cup.
    Differences in kettle and cup placement across configurations induce distinct role assignments for pouring and supporting.
\end{enumerate}

\section{Simulation Task Gallery}

\begin{figure}[h]
\centering
\includegraphics[width=\linewidth]{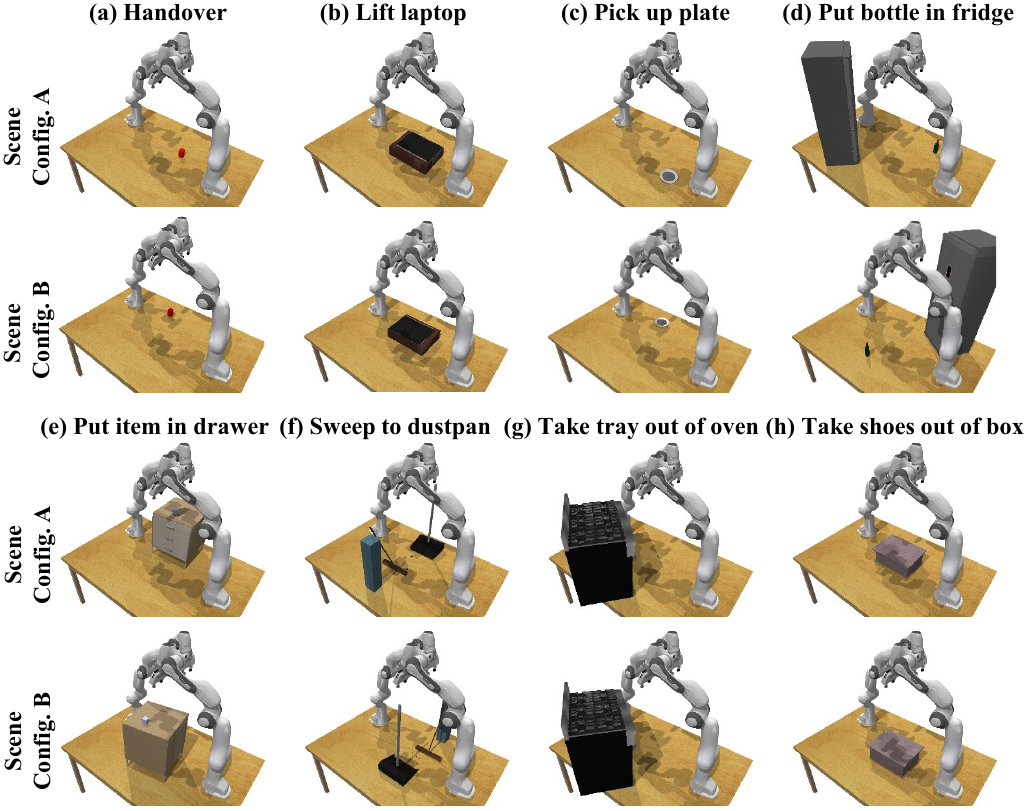}
\caption{
\textbf{Simulation task gallery.}
We show eight simulated bimanual tasks under two scene configurations. The configurations preserve the task goal and language instruction while varying scene layout or object accessibility, inducing different arm-role assignments.
}
\label{fig:supp_sim_data}
\end{figure}

As illustrated in Figure~\ref{fig:supp_sim_data}, the benchmark comprises eight bimanual tasks.
For each family, we design two distinct scene configurations, denoted Scene Config. A and Scene Config. B, resulting in a total of sixteen task variants.
The following description is intended to clarify the task construction and how the two configurations induce different arm-role assignments, rather than to define the language instruction used for training and evaluation.
Both configurations preserve the original task goal and language instruction but vary object layouts, target regions, tool placements, or articulated-object orientations to induce different functional role assignments between the two arms.
In all cases, the policy must infer the appropriate arm-role assignment from the scene without access to explicit role labels.

The following paragraphs provide additional clarification on the task procedure and role variations.

\begin{enumerate}
    \item \textbf{Handover.}
    The task involves transferring an object between the two arms.
    One arm grasps the object, while the other receives it to complete the handover.
    Compared with the original task, we expand the object initialization range.
    In Scene Config. A, the object is initialized closer to one side of the workspace; in Scene Config. B, it is initialized closer to the opposite side.
    The two regions partially overlap near the center, preventing the policy from relying on a fixed arm identity and requiring it to decide which arm should initiate the grasp based on the current scene.

    \item \textbf{Lift laptop.}
    The robot must lift a laptop from the table.
    One arm first pushes and adjusts the laptop into a graspable pose, while the other arm grasps and lifts it.
    The laptop position is randomized analogously to the handover task.
    The functional role assignment depends on which arm has better access to push the laptop and which arm is better positioned to grasp and lift it.

    \item \textbf{Pick up plate.}
    The robot needs to pick up a plate from the table.
    One arm presses the plate edge to create clearance, while the other arm grasps and lifts it.
    Plate positions are randomized across the workspace.
    The policy must assign roles according to which arm can more conveniently press the plate edge and which arm can more reliably grasp and lift the plate.

    \item \textbf{Put bottle in the fridge.}
    The scene contains a bottle and a fridge.
    One arm opens the fridge door, while the other arm grasps the bottle and places it inside.
    Scene configurations vary the relative accessibility of the bottle and fridge door, requiring the policy to infer which arm should operate the articulated door and which arm should manipulate the bottle.

    \item \textbf{Put item in drawer.}
    The scene contains a small item and a drawer.
    One arm opens the target drawer, while the other arm grasps the item and places it inside.
    We increase drawer-orientation randomization by allowing additional rotation.
    This changes which arm can more conveniently open the drawer, so the policy must adapt the arm-role assignment according to the drawer pose and object layout.

    \item \textbf{Sweep to dustpan.}
    The scene contains a broom, a dustpan, and target dirt.
    One arm holds and moves the broom to sweep the dirt, while the other arm holds the dustpan to collect it.
    Tool placements are randomized so that either arm may be closer to the broom or the dustpan.
    The policy must therefore assign the sweeping and supporting roles based on current tool locations rather than fixed arm identities.

    \item \textbf{Take the tray out of oven.}
    The scene contains an oven and a tray inside it.
    One arm opens the oven door, while the other arm reaches into the oven and takes out the tray.
    Both arms may operate the oven door depending on the scene configuration.
    This requires the policy to choose which arm should open the door and which arm should extract the tray based on accessibility.

    \item \textbf{Take shoes out of box.}
    The scene contains a hinged shoebox and a pair of shoes inside it.
    One arm opens the shoebox lid, while the other arm takes the shoes out of the box.
    The shoebox position is randomized similarly to the handover task.
    Because the box has a hinged structure, the easier opening direction depends on its pose and location; the policy must infer which arm should open the lid and assign the other arm to retrieve the shoes.
\end{enumerate}

\section{Evaluation Metrics}

We use task success rate as the primary evaluation metric, measured as the percentage of successful rollouts over held-out evaluation episodes. For each task, the evaluation set is balanced across role configurations, with 50 held-out episodes in the base-role configuration and 50 held-out episodes in the role-reversed configuration. Each episode is generated from randomized initial conditions. To ensure a fair comparison, all methods are evaluated on the same held-out episodes under the same evaluation protocol. This balanced evaluation design is independent of the training data ratio, e.g., (50{:}50) or (95{:}5), which only specifies the composition of the training demonstrations.

Beyond average task performance, we evaluate whether a policy performs consistently under the two role configurations. This is important because a policy may obtain a high overall success rate while still favoring one role configuration.

For each task \(t_i\), we evaluate the policy under both the base-role and role-reversed configurations. Let \(S_i^{\mathrm{base}}\) and \(S_i^{\mathrm{rev}}\) denote the success rates of task \(t_i\) under these two configurations, respectively, and let \(N\) denote the total number of tasks. All metrics are computed separately for each training data-ratio setting.

\begin{itemize}
\item \textbf{Configuration-level success.}
We report the average success rate for each role configuration:
\begin{equation}
\mathrm{Base}=\frac{1}{N}\sum_{i=1}^{N}S_i^{\mathrm{base}},
\qquad
\mathrm{Rev.}=\frac{1}{N}\sum_{i=1}^{N}S_i^{\mathrm{rev}}.
\end{equation}
These metrics indicate whether a method achieves comparable performance under the base-role and role-reversed configurations.

\item \textbf{Mean.}
The overall mean success rate averages performance over the two role configurations for each task:
\begin{equation}
\mathrm{Mean}=\frac{1}{N}\sum_{i=1}^{N}
\frac{S_i^{\mathrm{base}}+S_i^{\mathrm{rev}}}{2}.
\end{equation}
This metric summarizes overall performance across all evaluated role configurations.

\item \textbf{Harmonic mean.}
To emphasize balanced performance within each task, we report the harmonic mean between the two role configurations:
\begin{equation}
\mathrm{HM}=\frac{1}{N}\sum_{i=1}^{N}
\frac{2S_i^{\mathrm{base}}S_i^{\mathrm{rev}}}
{S_i^{\mathrm{base}}+S_i^{\mathrm{rev}}}.
\end{equation}
If \(S_i^{\mathrm{base}}+S_i^{\mathrm{rev}}=0\), the corresponding task-level harmonic mean is defined as zero.

\item \textbf{Worst-configuration performance.}
To measure performance under the weaker role configuration of each task, we compute
\begin{equation}
\mathrm{Worst}=\frac{1}{N}\sum_{i=1}^{N}
\min\left(S_i^{\mathrm{base}}, S_i^{\mathrm{rev}}\right).
\end{equation}
This metric reflects the average success rate of the lower-performing role configuration across tasks.

\item \textbf{Configuration gap.}
Finally, we measure the average absolute performance gap between the two role configurations:
\begin{equation}
\mathrm{Gap}=\frac{1}{N}\sum_{i=1}^{N}
\left|S_i^{\mathrm{base}}-S_i^{\mathrm{rev}}\right|.
\end{equation}
A smaller gap indicates more consistent performance between the base-role and role-reversed configurations. Since a small gap can also occur when both configurations have uniformly low success rates, we interpret Gap together with Mean, HM, and Worst rather than as a standalone measure of performance.

\end{itemize}

\section{Detailed Simulation Results}
\label{supp_sec:results}

\newcommand{\best}[1]{\textbf{#1}}
\newcommand{\second}[1]{\underline{#1}}

\begin{table*}[htbp]
\centering
\caption{
Detailed per-task simulation success rates under balanced and imbalanced demonstration compositions. Each entry reports success rates on the base-role and role-reversed configurations, shown as base / reversed. For the 95:5 setting, the reversed configuration corresponds to the underrepresented role configuration. \textbf{Bold} and \underline{underline} indicate the best and second-best results within each training composition and configuration.
PerAct$^2$ is single-task trained following its original protocol and is excluded from rank highlighting.
}
\vspace{2mm}
\label{tab:supp_main_results}
\scriptsize
\setlength{\tabcolsep}{3.0pt}
\renewcommand{\arraystretch}{1.08}

\resizebox{\textwidth}{!}{
\begin{tabular}{@{}lccccccccc@{}}
\toprule
\rowcolor{gray!12}
\multicolumn{10}{c}{\textbf{Balanced demonstrations: 50:50}} \\
\midrule
\multirow{2}{*}{Method}
& \multicolumn{8}{c}{Task Success Rate (\%, base / reversed)}
& \multirow{2}{*}{\makecell{Config.\\Avg.}} \\
\cmidrule(lr){2-9}
& Handover
& \makecell{Lift\\laptop}
& \makecell{Pick up\\plate}
& \makecell{Put bottle\\in fridge}
& \makecell{Put item\\in drawer}
& \makecell{Sweep to\\dustpan}
& \makecell{Take tray\\out of oven}
& \makecell{Take shoes\\out of box}
& \\
\midrule

DP3
& 34 / 34
& 0 / 0
& 18 / 6
& 0 / 0
& 0 / 0
& 0 / 0
& 0 / 0
& 42 / 58
& 11.75 / 12.25 \\

ACT
& 34 / 22
& 20 / 0
& 6 / 6
& 38 / \second{18}
& 0 / 2
& 10 / 8
& 0 / 2
& 54 / \second{96}
& 20.25 / 19.25 \\

PerAct$^2$
& 10 / 28
& 60 / 20
& 68 / 60
& 0 / 2
& 36 / 42
& 0 / 12
& 2 / 6
& 0 / 0
& 22.00 / 21.25 \\

\cmidrule(lr){1-10}

$\pi_0$
& 52 / 62
& 24 / \second{26}
& 6 / 6
& 2 / 0
& 4 / 6
& 2 / 4
& 16 / 14
& 44 / 74
& 18.75 / 24.00 \\

\rowcolor{gray!8}
$\pi_0$ + BiRoAD
& 66 / 70
& 20 / 22
& 8 / 10
& 8 / 4
& 12 / \second{12}
& 6 / 10
& 24 / 18
& 64 / 72
& 26.00 / 27.25 \\

\cmidrule(lr){1-10}

3DFA
& \second{86} / \second{94}
& \best{78} / 14
& \second{64} / \second{74}
& \second{88} / 8
& \second{82} / \best{96}
& \second{46} / \best{68}
& \second{74} / \best{86}
& \second{78} / \best{100}
& \second{74.50} / \second{67.50} \\

\rowcolor{gray!8}
\textbf{3DFA + BiRoAD}
& \best{92} / \best{98}
& \second{76} / \best{92}
& \best{84} / \best{82}
& \best{92} / \best{46}
& \best{90} / \best{96}
& \best{50} / \second{52}
& \best{80} / \second{76}
& \best{100} / 60
& \best{83.00} / \best{75.25} \\

\bottomrule
\end{tabular}
}

\vspace{0.45em}

\resizebox{\textwidth}{!}{
\begin{tabular}{@{}lccccccccc@{}}
\toprule
\rowcolor{gray!12}
\multicolumn{10}{c}{\textbf{Imbalanced demonstrations: 95:5, reversed role underrepresented}} \\
\midrule
\multirow{2}{*}{Method}
& \multicolumn{8}{c}{Task Success Rate (\%, base / reversed)}
& \multirow{2}{*}{\makecell{Config.\\Avg.}} \\
\cmidrule(lr){2-9}
& Handover
& \makecell{Lift\\laptop}
& \makecell{Pick up\\plate}
& \makecell{Put bottle\\in fridge}
& \makecell{Put item\\in drawer}
& \makecell{Sweep to\\dustpan}
& \makecell{Take tray\\out of oven}
& \makecell{Take shoes\\out of box}
& \\
\midrule

DP3
& 40 / 46
& 0 / 0
& 38 / 0
& 0 / 0
& 0 / 0
& 0 / 0
& 0 / 0
& 18 / 2
& 12.00 / 6.00 \\

ACT
& 38 / 18
& 54 / 0
& 14 / 2
& 0 / 0
& 30 / 0
& 22 / 0
& 32 / 30
& 72 / 0
& 32.75 / 6.25 \\

PerAct$^2$
& 8 / 22
& 6 / 22
& 78 / 78
& 0 / 2
& 10 / 0
& 0 / 24
& 4 / 10
& 0 / 0
& 13.25 / 19.75 \\

\cmidrule(lr){1-10}

$\pi_0$
& 70 / 50
& 28 / 2
& 26 / 14
& 2 / 0
& 8 / 0
& 2 / 0
& 28 / 36
& 22 / 10
& 23.25 / 14.00 \\

\rowcolor{gray!8}
$\pi_0$ + BiRoAD
& 76 / 56
& 36 / \second{22}
& 14 / 14
& 10 / 0
& 14 / 0
& 26 / 0
& 42 / 34
& 76 / 16
& 36.75 / 17.75 \\

\cmidrule(lr){1-10}

3DFA
& \best{100} / \second{84}
& \best{100} / 20
& \second{64} / \second{44}
& \second{80} / \best{22}
& \second{90} / \second{24}
& \second{50} / \best{74}
& \second{88} / \best{90}
& \second{78} / \second{42}
& \second{81.25} / \second{50.00} \\

\rowcolor{gray!8}
\textbf{3DFA + BiRoAD}
& \second{96} / \best{86}
& \second{96} / \best{78}
& \best{72} / \best{66}
& \best{92} / \best{22}
& \best{94} / \best{42}
& \best{86} / \second{32}
& \best{94} / \second{86}
& \best{94} / \best{94}
& \best{90.50} / \best{63.25} \\

\bottomrule
\end{tabular}
}
\label{tab:supp_per_task_results}

\end{table*}

We provide the detailed per-task simulation results corresponding to the main results in the paper.
Table~\ref{tab:supp_per_task_results} reports task success rates for each method under the balanced and imbalanced demonstration compositions.
Each entry is shown as the success rate on the base-role configuration and the role-reversed configuration.

For fair comparison, all methods use the same data splits, observation resolution, action representation, language setting, and evaluation protocol.
Except for PerAct$^2$, which follows its original single-task training protocol, all other methods are trained in a multitask setting.
Accordingly, ranking highlights in Table~\ref{tab:supp_per_task_results} are applied only among multitask-trained methods.

These results complement the aggregated metrics in the main paper by showing how each method performs across individual task families and role configurations.

\section{Additional Experiments}
\label{supp_sec:additional_experiments}

\subsection{Data-Level Baselines}

We further compare BiRoAD with two data-level strategies for addressing role imbalance on 3DFA under the 95:5 training split: arm-swap augmentation and minority oversampling.

\paragraph{Arm-swap augmentation.}
We augment the training data by swapping the left- and right-arm proprioceptive inputs, target actions, and corresponding wrist-camera observations with probability 0.5. This baseline examines whether exposure to swapped arm-specific signals can reduce bias toward fixed arm identities.

\paragraph{Minority oversampling.}
We rebalance training by oversampling the minority role-reversed demonstrations, giving the two role configurations equal expected sampling frequency. This baseline examines the effect of increasing training exposure to the underrepresented role configuration.

\begin{table}[htbp]
\centering
\caption{
Comparison with data-level baselines on 3DFA under the 95:5 training split.
}
\vspace{-1mm}
\small
\setlength{\tabcolsep}{2pt}
\label{tab:supp_additional_experiments}
\resizebox{0.95\textwidth}{!}{
\begin{tabular*}{\linewidth}{@{\extracolsep{\fill}}lccccc@{}}
\toprule
Method
& Config.-level Avg. $\uparrow$
& Mean $\uparrow$
& HM $\uparrow$
& Worst $\uparrow$
& Gap $\downarrow$ \\
\midrule
3DFA
& \twoval{81.25}{50.00}
& 65.62 & 56.56 & 46.75 & 37.75 \\
+ arm-swap augmentation
& \twoval{86.50}{57.25}
& 71.88 & 64.48 & 55.75 & 32.25 \\
+ minority oversampling
& \twoval{78.25}{55.75}
& 67.00 & 63.15 & 55.25 & \textbf{23.50} \\
+ \biroad{}
& \textbf{\twoval{90.50}{63.25}}
& \textbf{76.88} & \textbf{71.21} & \textbf{63.25} & 27.25 \\
\bottomrule
\end{tabular*}
}
\vspace{-1mm}
\end{table}

\paragraph{Results.}
Table~\ref{tab:supp_additional_experiments} shows that arm-swap augmentation improves performance on both role configurations, while minority oversampling improves the underrepresented configuration at a small cost to majority performance. BiRoAD achieves the strongest overall performance, highlighting the benefit of explicitly structuring shared and role-specific representations for role adaptation under imbalance.

\subsection{Branch Representation Analysis}
\label{supp_sec:branch_representation}

We further examine whether the symmetric and antisymmetric branches exhibit the intended complementary structure, and how strongly they contribute during inference.

\paragraph{Linear probes.}
We train task-specific linear probes on frozen projected symmetric and antisymmetric features to predict the functional role of each arm. The antisymmetric features achieve task-averaged accuracies of 98.8\% and 98.1\% under the 50:50 and 95:5 settings, respectively, while the symmetric features yield 50.0\% in both settings. These results show that functional-role information is linearly accessible from the antisymmetric branch, consistent with its intended role-specific function.

\paragraph{Branch activation.}
As shown in Table~\ref{tab:supp_branch_activation}, both branches exhibit non-negligible activation magnitudes during inference, approximately 10--20\% of the input-feature magnitude. Both $\alpha_S$ and $\alpha_A$ increase under the imbalanced 95:5 setting compared with the balanced 50:50 setting, suggesting greater reliance on BiRoAD under role imbalance. Within the 95:5 setting, $\alpha_A$ is higher for the minority role-reversed configuration than for the base-role configuration ($0.182$ vs.\ $0.144$), indicating a stronger role-specific contribution for the underrepresented configuration.

\begin{table}[htbp]
\centering
\small
\caption{
Relative branch activation magnitudes for 3DFA + \biroad{}.
$\alpha_A$ and $\alpha_S$ denote the projected antisymmetric and symmetric contributions relative to the input-feature magnitude.
}
\label{tab:supp_branch_activation}
\begin{tabular}{@{}lcc|cc@{}}
\toprule
\multirow{2}{*}{Configuration}
& \multicolumn{2}{c}{50:50}
& \multicolumn{2}{c}{95:5} \\
\cmidrule(lr){2-3}
\cmidrule(lr){4-5}
& $\alpha_A$ & $\alpha_S$
& $\alpha_A$ & $\alpha_S$ \\
\midrule
Base-role
& 0.108 & 0.127
& 0.144 & 0.196 \\
Role-reversed
& 0.113 & 0.126
& 0.182 & 0.184 \\
\bottomrule
\end{tabular}
\end{table}

\section{Training and Inference Cost}

We report the training and inference costs of the base policies and their BiRoAD-enhanced variants.
For the 3DFA backbone, both the base policy and the BiRoAD-enhanced policy are trained from scratch.
The 3DFA base policy was trained on 2 A100 GPUs for 500k iterations, requiring approximately 29 hours, with a peak memory usage of 5.54 GB per GPU.
Under the same GPU configuration and number of training iterations, the 3DFA + BiRoAD model required approximately 30 hours, with a peak memory usage of 5.56 GB per GPU.

For the $\pi_0$ backbone, both the base policy and the BiRoAD-enhanced policy are fully fine-tuned from the same pretrained checkpoint for 50k iterations on 4 A100 GPUs. Reported results use the best-performing checkpoints within this budget, selected at 30k iterations for the base policy and 50k for $\pi_0$ + BiRoAD.
The $\pi_0$ base policy required approximately 19 hours for the full 50k iterations, with a peak memory usage of 32.7 GB per GPU.
Under the same GPU configuration and training budget, the $\pi_0$ + BiRoAD model required approximately 19.5 hours, with a peak memory usage of 32.6 GB per GPU.

During inference, BiRoAD adds only a lightweight symmetric--antisymmetric decomposition to the bimanual trajectory or action-token representations, while leaving the backbone encoders and the denoising schedule unchanged. As a result, its inference latency remains comparable to that of the base policy.

\end{document}